\documentclass[conference, a4paper]{IEEEtran}
\IEEEoverridecommandlockouts

\usepackage{cite}
\usepackage{amsmath,amssymb,amsfonts}
\usepackage{algorithmic}
\usepackage{graphicx}
\usepackage{textcomp}
\usepackage{hyperref}
\usepackage[svgnames]{xcolor}
\usepackage{enumitem}

\usepackage{colortbl}
\usepackage{booktabs}
\usepackage{multirow}
\usepackage{array}
\usepackage{tabularx}
\usepackage{makecell}
\usepackage{siunitx}
\usepackage[font=footnotesize]{subcaption}

\def\BibTeX{{\rm B\kern-.05em{\sc i\kern-.025em b}\kern-.08em
		T\kern-.1667em\lower.7ex\hbox{E}\kern-.125emX}}
\begin{document}
	
	\title{Analysis of Pseudo-Labeling for Online Source-Free Universal Domain Adaptation}
	
	\author{
		\IEEEauthorblockN{Pascal Schlachter, Jonathan Fuss, Bin Yang}
		\IEEEauthorblockA{\textit{Institute of Signal Processing and System Theory}, University of Stuttgart, Germany\\
			\{pascal.schlachter, bin.yang\}@iss.uni-stuttgart.de}
	}
	
	\maketitle
	
	\begin{abstract}
		A domain (distribution) shift between training and test data often hinders the real-world performance of deep neural networks, necessitating unsupervised domain adaptation (UDA) to bridge this gap. Online source-free UDA has emerged as a solution for practical scenarios where access to source data is restricted and target data is received as a continuous stream. However, the open-world nature of many real-world applications additionally introduces category shifts meaning that the source and target label spaces may differ. Online source-free universal domain adaptation (SF-UniDA) addresses this challenge. Existing methods mainly rely on self-training with pseudo-labels, yet the relationship between pseudo-labeling and adaptation outcomes has not been studied yet. To bridge this gap, we conduct a systematic analysis through controlled experiments with simulated pseudo-labeling, offering valuable insights into pseudo-labeling for online SF-UniDA. Our findings reveal a substantial gap between the current state-of-the-art and the upper bound of adaptation achieved with perfect pseudo-labeling. Moreover, we show that a contrastive loss enables effective adaptation even with moderate pseudo-label accuracy, while a cross-entropy (CE) loss, though less robust to pseudo-label errors, achieves superior results when pseudo-labeling approaches perfection. Lastly, our findings indicate that pseudo-label accuracy is in general more crucial than quantity, suggesting that prioritizing fewer but high-confidence pseudo-labels is beneficial. Overall, our study highlights the critical role of pseudo-labeling in (online) SF-UniDA and provides actionable insights to drive future advancements in the field. Our code is available at \url{https://github.com/pascalschlachter/PLAnalysis}.
	\end{abstract}
	
	
	\section{Introduction}
	Distribution shifts between training and test data often limit the real-world performance of deep neural networks. Accordingly, unsupervised domain adaptation (UDA) is necessary to adapt a pre-trained source model to unlabeled target data and, in this way, enables a robust performance across varying conditions. While early domain adaptation methods assume full access to source data, this is often impractical due to privacy concerns or memory constraints. Consequently, source-free UDA emerged, relying solely on the pre-trained source model and unlabeled target data for adaptation. This is often also referred to as test-time adaptation.
	
	Considering the open-world nature of many real-world settings, recent advancements in source-free UDA aim to handle category (label) shifts alongside domain shifts. Specifically, this refers to scenarios where the source label space, $\mathcal{Y}_s$, and the target label space, $\mathcal{Y}_t$, are different ($\mathcal{Y}_s\neq\mathcal{Y}_t$). There are three cases: partial-set domain adaptation (PDA) ($\mathcal{Y}_t\subset \mathcal{Y}_s$), open-set domain adaptation (ODA) ($\mathcal{Y}_s\subset \mathcal{Y}_t$) and open-partial-set domain adaptation (OPDA) ($\mathcal{Y}_s\cap\mathcal{Y}_t\neq\emptyset$, $\mathcal{Y}_s\nsubseteq\mathcal{Y}_t$, $\mathcal{Y}_s\nsupseteq\mathcal{Y}_t$). Since prior knowledge of the category shift is typically unavailable, source-free universal domain adaptation (SF-UniDA) \cite{GLC} aims to perform well across all three category shifts by detecting samples of new classes as unknown while accurately classifying samples of known classes.
	
	Moreover, in many real-world applications, target data is received as a continuous stream rather than a fixed dataset with unrestricted access. To enable real-time processing of such data streams, adaptation and inference must be carried out simultaneously, with each test batch processed only once and handled sequentially. This setup is referred to as online UDA.
	
	All existing methods towards SF-UniDA, both online \cite{COMET, GMM} and offline \cite{GLC, GLC++, LEAD}, rely on self-training with pseudo-labels. Thereby, the robustness of pseudo-labeling intuitively plays a crucial role in the success of the adaptation. However, the underlying relationship between pseudo-labeling and adaptation outcomes has not been studied yet. In this paper, we aim to address this gap and provide an extensive analysis of pseudo-labeling for online SF-UniDA. Specifically, we seek to answer the following research questions:
	\begin{enumerate}[label=Q\arabic*., leftmargin=*]
		\item What is the upper bound of adaptation that would be achieved by a perfect pseudo-labeling and how far away is the current state-of-the-art from this upper bound?
		\item What are the minimum requirements that pseudo-labeling must meet for adaptation to be effective, meaning that the performance improves w.r.t. the unadapted source model?
		\item Existing SF-UniDA methods primarily rely on two types of loss functions: contrastive loss \cite{COMET, GMM} and cross-entropy (CE) loss \cite{GLC, GLC++, LEAD}. How do they differ, particularly in terms of robustness to erroneous pseudo\mbox{-}labels?
		\item Following \cite{feng2021open}, we exclude samples with uncertain pseudo-labels from the adaptation in our recent online SF-UniDA methods \cite{COMET, GMM}. The underlying intuition is that pseudo-label quality (accuracy) is more important than the quantity of samples used for adaptation. Does empirical evidence validate this hypothesis in practice?
	\end{enumerate}
	
	\section{Related Work}
	%
	To date, the only universal domain adaptation (UniDA) methods that strictly adhere to the "source-free" definition given by \cite{GLC}---meaning they rely solely on a standard pre-trained source model as the only source knowledge---are \cite{GLC, GLC++, LEAD, COMET, GMM}. While other UniDA approaches \cite{universal_sf_da, kundu2020towards, umad, coca} also claim to be source-free because they do not access source data during adaptation, they impose specific training procedures or architectural constraints that differentiate them from strictly source-free methods. This limits their practical applicability.
	
	Source-free UniDA (SF-UniDA) was first introduced by \cite{GLC} with the Global and Local Clustering (GLC) method, which employs clustering-based pseudo-labeling. Adaptation is performed using a combination of CE loss and a kNN-based loss function. \cite{GLC++} extends this approach by incorporating an additional contrastive loss to enhance adaptation. Building on this foundation, \cite{LEAD} introduces Learning Decomposition (LEAD), a novel pseudo-labeling technique that separates features into source-known and -unknown components. This especially enables the use of individual rejection thresholds for unknown class identification during pseudo-labeling and allows the integration of confidence-weighting into the CE loss. Despite these variations, all these methods rely on entropy-based rejection for unknown classes during inference.
	
	Recently, we demonstrated that these methods designed for offline SF-UniDA do not generalize well to the online setting because single batches fail to sufficiently capture the underlying data distribution \cite{COMET, GMM}. To address this limitation, we proposed two approaches specifically tailored for online SF-UniDA. Both methods follow similar adaptation strategies, combining a contrastive loss with an additional loss that enhances entropy-based distinguishability between known and unknown classes. However, they differ in their pseudo-labeling approaches: \cite{COMET} employs a mean teacher, whereas \cite{GMM} uses a Gaussian mixture model (GMM) in the feature space.
	

	\section{Simulation}
	\subsection{Setup}
	\subsubsection{Pseudo-Label Simulation}
	Pseudo-labeling is defined by two key metrics: pseudo-label accuracy and the proportion of samples used for adaptation, which we refer to as pseudo-label quality and quantity, respectively. To systematically investigate their impact on adaptation performance, we require a simulation that mimics real-world pseudo-labeling while enabling precise control over both factors. To achieve this, we first perform a forward pass without adaptation and compute the normalized entropy of the source model's predictions $\boldsymbol{p}_i$:
	\begin{align}
		I(\boldsymbol{p}_i)=-\frac{1}{\log |\mathcal{Y}_s|}\cdot\boldsymbol{p}_i^T\cdot\log \boldsymbol{p}_i
		\label{eq:entropy}
	\end{align}
	where $|\mathcal{Y}_s|$ denotes the number of known classes.
	
	To achieve a desired pseudo-label quantity by selecting samples for adaptation, we mimic the idea of \cite{feng2021open}, \cite{COMET}, and \cite{GMM}, which suggests that low-entropy samples indicate confident predictions of known classes, while high-entropy samples can be reliably classified as unknown. However, predictions with mid-range entropy values cannot be assigned to a class with confidence and therefore the corresponding samples are excluded from adaptation. To formalize this selection, we compute a confidence score based on the distance of each sample’s entropy to the nearest extreme (i.e., 0 or 1):
	\begin{align}
		d_i = \min(I(\boldsymbol{p}_i), 1 - I(\boldsymbol{p}_i))
	\end{align}
	A certain pseudo-label quantity $q$ is then achieved by selecting the top $q\%$ of samples with the smallest entropy distances $d_i$.
	
	Next, to control the pseudo-label quality $a$, the selected samples are divided into two groups: those that receive correct pseudo-labels and those that are deliberately mislabeled. To maintain a realistic pseudo-labeling process, we again use the entropy distance $d_i$ as an indicator. Samples with lower distances are more likely to be pseudo-labeled correctly while those with higher distances are more likely to be mislabeled. Specifically, among the selected samples, the top $a\%$ with the smallest distances $d_i$ are assigned their correct pseudo-labels, while the remaining $(1-a)\%$ receive incorrect pseudo-labels.
	
	Assigning correct pseudo-labels is straightforward, as the ground truth labels are provided by the dataset. However, assigning incorrect pseudo-labels requires careful consideration to ensure a realistic simulation of pseudo-labeling errors. Concretely, a realistic pseudo-labeling process reflects the model’s natural tendency to confuse similar classes, meaning that misclassifications should follow the patterns observed in real predictions rather than being assigned arbitrarily. To achieve this, we assign incorrect pseudo-labels based on the model’s predicted probabilities. If the true label corresponds to an unknown class, the sample is pseudo-labeled as the known class with the highest predicted probability. If the true label belongs to a known class, the sample is pseudo-labeled as the class with the highest predicted probability, excluding the correct label. However, if this probability falls below a threshold $\tau_i$, the sample is instead pseudo-labeled as unknown.
	
	To ensure that samples with higher prediction uncertainties are more likely to be pseudo-labeled as unknown, we define this probability threshold $\tau_i$ as an adaptive threshold that dynamically increases with entropy:
	\begin{align}
		\tau_i = \alpha \cdot I(\boldsymbol{p}_i)
	\end{align}
	where $\alpha$ is a scaling factor and $I(\boldsymbol{p}_i)$ is the normalized entropy. Our analysis of a mean teacher \cite{COMET} and GMM-based \cite{GMM} pseudo-labeling revealed that when samples from known classes are incorrectly pseudo-labeled, they are predominantly assigned to the unknown class. To replicate this behavior, we empirically found that setting $\alpha=1$, i.e. $\tau_i=I(\boldsymbol{p}_i)$, effectively mimics this tendency.
	
	\subsubsection{Loss Functions}
	Next, the generated pseudo-labels are used to guide the adaptation process. In the literature, SF-UniDA methods primarily employ two types of loss functions, which we aim to compare in this work.
	
	The first approach combines a contrastive loss to encourage meaningful feature extraction with an entropy- or KL divergence-based loss to enforce separation between classifier outputs for known and unknown class samples \cite{COMET, GMM}. Here, we adopt the same loss function as COMET \cite{COMET}. To maintain a strictly source-free setting, we avoid using source prototypes and instead compute target-domain prototypes using a class-wise exponential moving average based on the pseudo-labels.
	
	The second approach primarily relies on a CE loss, where the unknown class is represented by uniformly distributed label vectors \cite{GLC, GLC++, LEAD}. Although these works combine CE with additional loss functions, we opt to use the CE loss alone in this study, as the other losses are either unsuitable for online settings---such as a kNN-based loss---or specifically designed to support particular pseudo-labeling strategies.
	
	\subsubsection{Dataset}
	For our experiments, we use the widely adopted public domain adaptation dataset VisDA-C \cite{visda}, which covers a 12-class classification problem\footnote{The classes are airplane, bicycle, bus, car, horse, knife, motorcycle, person, plant, skateboard, train, and truck.}. Its source domain contains 152,397 synthetically generated renderings of 3D objects while the target domain consists of 55,388 real-world images taken from the Microsoft COCO dataset \cite{coco}. To create category shift scenarios, we divide the 12 classes into shared classes (present in both domains), source-private classes, and target-private classes as follows:
	\begin{itemize}
		\item PDA: 6 shared, 6 source-private, 0 target-private
		\item ODA: 6 shared, 0 source-private, 6 target-private
		\item OPDA: 6 shared, 3 source-private, 3 target-private
	\end{itemize}
	
	\subsubsection{Implementation details}
	We use the same experimental framework as \cite{COMET}. The source model is based on a ResNet-50 \cite{resnet} backbone and pre-trained using a standard supervised procedure. For domain adaptation, we use a batch size of 64 and optimize with SGD, applying a momentum of 0.9 and a learning rate of $0.001$. During inference, we set a fixed entropy threshold of $0.5$ to reject samples as unknown. Regarding the hyperparameters of the contrastive loss-based approach, we set the temperature to $0.1$, the dimensions of the projection space to $128$ and use $\lambda=0.01$ as loss balancing factor.
	
	We evaluate PDA using standard accuracy, while for ODA and OPDA, we report the H-score:
	\begin{align}
		\text{H-score} = \frac{2 \cdot acc_k \cdot acc_u}{acc_k + acc_u}
	\end{align}
	which is the harmonic mean of the known and unknown class accuracies, \( acc_k \) and \( acc_u \). Each experiment is repeated three times, and we report the average performance.

	\subsection{Results}
	The results are presented in the Tables \ref{tab:results_PDA}, \ref{tab:results_ODA}, and \ref{tab:results_OPDA}. To enhance clarity and facilitate interpretation, we have color-coded each cell. Cells representing results worse than the source-only baseline are highlighted in red. For results improving upon the source-only baseline, a new color is introduced every $10\%$, with dark green shades finally indicating improvements greater than $40\%$. The source-only performance serves as the baseline, with an accuracy of $17.1\%$ for PDA, an H-score of $31.7\%$ for ODA, and an H-score of $26.9\%$ for OPDA.

	\begin{table*}[t!]
		\centering
		\caption{Accuracies in $\%$ for the PDA scenario. The source-only baseline performance is $17.1\%$.}
		\label{tab:results_PDA}

		\setlength{\tabcolsep}{2.5pt}
		\renewcommand{\arraystretch}{1.2}
		
		\newcommand{\heatmap}[1]{%
			\begingroup
			\def\val{#1} 
			\ifdim\val pt<17.1pt \cellcolor{red!80}#1%
			\else\ifdim\val pt<27.1pt \cellcolor{orange!60}#1%
			\else\ifdim\val pt<37.1pt \cellcolor{yellow!60}#1%
			\else\ifdim\val pt<47.1pt \cellcolor{GreenYellow!80}#1%
			\else\ifdim\val pt<57.1pt \cellcolor{green!80}#1%
			\else \cellcolor{Green!80}#1%
			\fi\fi\fi\fi\fi
			\endgroup
		}
		
		\begin{subtable}[t]{0.49\textwidth}
			\caption{Contrastive loss}
			\label{tab:results_PDA_1}
			\begin{tabularx}{\linewidth}{p{0.25cm} | c |*{11}{>{\centering\arraybackslash}X}}
				\toprule
				\multirow{11}{*}[-8ex]{\rotatebox{90}{pseudo-label quantity in $\%$}} &  & \multicolumn{11}{c}{pseudo-label quality in $\%$} \\ 
				\cmidrule(lr){3-13}
				&  & {$0$} & $10$ & $20$ & $30$ & $40$ & $50$ & $60$ & $70$ & $80$ & $90$ & $100$\\
				\cmidrule(lr){3-13}
				& $10$ & \heatmap{1.9} & \heatmap{8.0} & \heatmap{10.4} & \heatmap{7.9} & \heatmap{28.5} & \heatmap{20.8} & \heatmap{34.0} & \heatmap{48.1} & \heatmap{49.2} & \heatmap{57.7} & \heatmap{63.7} \\
				
				& $20$ & \heatmap{0.9} & \heatmap{1.6} & \heatmap{9.9} & \heatmap{22.5} & \heatmap{27.6} & \heatmap{24.9} & \heatmap{45.4} & \heatmap{44.9} & \heatmap{51.2} & \heatmap{54.2} & \heatmap{71.8} \\
				
				& $30$ & \heatmap{0.7} & \heatmap{1.5} & \heatmap{12.0} & \heatmap{21.5} & \heatmap{23.4} & \heatmap{28.2} & \heatmap{37.4} & \heatmap{39.8} & \heatmap{41.3} & \heatmap{54.6} & \heatmap{74.3} \\
				
				& $40$ & \heatmap{7.0} & \heatmap{5.0} & \heatmap{10.4} & \heatmap{17.5} & \heatmap{23.9} & \heatmap{25.8} & \heatmap{35.0} & \heatmap{38.5} & \heatmap{43.5} & \heatmap{53.1} & \heatmap{74.6} \\
				
				& $50$ & \heatmap{2.9} & \heatmap{7.6} & \heatmap{11.8} & \heatmap{22.3} & \heatmap{24.4} & \heatmap{19.7} & \heatmap{29.6} & \heatmap{37.3} & \heatmap{45.4} & \heatmap{38.5} & \heatmap{74.4} \\
				
				& $60$ & \heatmap{2.0} & \heatmap{6.6} & \heatmap{7.0} & \heatmap{19.1} & \heatmap{19.9} & \heatmap{21.0} & \heatmap{26.9} & \heatmap{26.8} & \heatmap{29.0} & \heatmap{36.1} & \heatmap{73.6} \\
				
				& $70$ & \heatmap{3.8} & \heatmap{6.9} & \heatmap{9.6} & \heatmap{18.6} & \heatmap{21.3} & \heatmap{19.4} & \heatmap{26.8} & \heatmap{25.2} & \heatmap{31.6} & \heatmap{44.5} & \heatmap{75.0} \\
				
				& $80$ & \heatmap{1.6} & \heatmap{11.0} & \heatmap{11.5} & \heatmap{19.8} & \heatmap{18.3} & \heatmap{26.4} & \heatmap{24.4} & \heatmap{27.2} & \heatmap{26.8} & \heatmap{35.9} & \heatmap{74.7} \\
				
				& $90$ & \heatmap{4.6} & \heatmap{8.3} & \heatmap{10.8} & \heatmap{17.2} & \heatmap{18.5} & \heatmap{17.7} & \heatmap{21.6} & \heatmap{22.7} & \heatmap{29.5} & \heatmap{41.2} & \heatmap{74.8} \\
				
				& $100$ & \heatmap{4.4} & \heatmap{6.2} & \heatmap{17.8} & \heatmap{17.9} & \heatmap{18.1} & \heatmap{19.0} & \heatmap{18.4} & \heatmap{23.4} & \heatmap{29.0} & \heatmap{33.1} & \heatmap{75.6} \\
				
				\bottomrule
			\end{tabularx}
		\end{subtable}\hfill
		\begin{subtable}[t]{0.49\textwidth}
			\caption{Cross-entropy loss}
			\label{tab:results_PDA_2}
			\begin{tabularx}{\linewidth}{p{0.25cm} | c |*{11}{>{\centering\arraybackslash}X}}
				\toprule
				\multirow{11}{*}[-8ex]{\rotatebox{90}{pseudo-label quantity in $\%$}} &  & \multicolumn{11}{c}{pseudo-label quality in $\%$} \\ 
				\cmidrule(lr){3-13}
				&  & $0$ & $10$ & $20$ & $30$ & $40$ & $50$ & $60$ & $70$ & $80$ & $90$ & $100$\\
				\cmidrule(lr){3-13}
				& $10$ & \heatmap{0.1} & \heatmap{0.2} & \heatmap{0.5} & \heatmap{0.6} & \heatmap{1.8} & \heatmap{4.7} & \heatmap{14.6} & \heatmap{31.0} & \heatmap{48.2} & \heatmap{70.9} & \heatmap{83.0} \\
				
				& $20$ & \heatmap{0.1} & \heatmap{0.2} & \heatmap{0.2} & \heatmap{0.5} & \heatmap{1.4} & \heatmap{5.3} & \heatmap{14.8} & \heatmap{34.4} & \heatmap{57.5} & \heatmap{78.1} & \heatmap{89.2} \\
				
				& $30$ & \heatmap{0.1} & \heatmap{0.2} & \heatmap{0.3} & \heatmap{0.4} & \heatmap{1.2} & \heatmap{4.7} & \heatmap{14.6} & \heatmap{33.0} & \heatmap{58.6} & \heatmap{79.9} & \heatmap{90.7} \\
				
				& $40$ & \heatmap{0.1} & \heatmap{0.1} & \heatmap{0.2} & \heatmap{0.4} & \heatmap{1.3} & \heatmap{4.6} & \heatmap{14.7} & \heatmap{33.7} & \heatmap{59.1} & \heatmap{80.9} & \heatmap{91.4} \\
				
				& $50$ & \heatmap{0.1} & \heatmap{0.1} & \heatmap{0.2} & \heatmap{0.3} & \heatmap{1.1} & \heatmap{4.8} & \heatmap{14.8} & \heatmap{34.1} & \heatmap{59.4} & \heatmap{81.1} & \heatmap{91.7} \\
				
				& $60$ & \heatmap{0.1} & \heatmap{0.1} & \heatmap{0.2} & \heatmap{0.4} & \heatmap{1.3} & \heatmap{5.2} & \heatmap{15.4} & \heatmap{33.7} & \heatmap{59.4} & \heatmap{81.5} & \heatmap{91.8} \\
				
				& $70$ & \heatmap{0.1} & \heatmap{0.1} & \heatmap{0.2} & \heatmap{0.4} & \heatmap{1.5} & \heatmap{5.4} & \heatmap{15.3} & \heatmap{33.4} & \heatmap{59.3} & \heatmap{81.2} & \heatmap{91.9} \\
				
				& $80$ & \heatmap{0.1} & \heatmap{0.1} & \heatmap{0.1} & \heatmap{0.4} & \heatmap{1.5} & \heatmap{5.6} & \heatmap{15.7} & \heatmap{34.0} & \heatmap{59.5} & \heatmap{81.6} & \heatmap{91.9} \\
				
				& $90$ & \heatmap{0.1} & \heatmap{0.1} & \heatmap{0.2} & \heatmap{0.4} & \heatmap{1.7} & \heatmap{6.2} & \heatmap{16.4} & \heatmap{34.6} & \heatmap{59.6} & \heatmap{81.7} & \heatmap{91.9} \\
				
				& $100$ & \heatmap{0.1} & \heatmap{0.1} & \heatmap{0.2} & \heatmap{0.5} & \heatmap{1.8} & \heatmap{6.5} & \heatmap{17.1} & \heatmap{35.1} & \heatmap{60.5} & \heatmap{82.0} & \heatmap{91.9} \\
				
				\bottomrule
			\end{tabularx}
		\end{subtable}
	\end{table*}
	
	\begin{table*}[t!]
		\centering
		\caption{H-scores in $\%$ for the ODA scenario. The source-only baseline performance is $31.7\%$.}
		\label{tab:results_ODA}

		\setlength{\tabcolsep}{2.5pt}
		\renewcommand{\arraystretch}{1.2}
		
		\newcommand{\heatmap}[1]{%
			\begingroup
			\def\val{#1} 
			\ifdim\val pt<31.7pt \cellcolor{red!80}#1%
			\else\ifdim\val pt<41.7pt \cellcolor{orange!60}#1%
			\else\ifdim\val pt<51.7pt \cellcolor{yellow!60}#1%
			\else\ifdim\val pt<61.7pt \cellcolor{GreenYellow!80}#1%
			\else\ifdim\val pt<71.7pt \cellcolor{green!80}#1%
			\else \cellcolor{Green!80}#1%
			\fi\fi\fi\fi\fi
			\endgroup
		}
		
		\begin{subtable}[t]{0.49\textwidth}
			\caption{Contrastive loss}
			\label{tab:results_ODA_1}
			\begin{tabularx}{\linewidth}{p{0.25cm} | c |*{11}{>{\centering\arraybackslash}X}}
				\toprule
				\multirow{11}{*}[-8ex]{\rotatebox{90}{pseudo-label quantity in $\%$}} &  & \multicolumn{11}{c}{pseudo-label quality in $\%$} \\ 
				\cmidrule(lr){3-13}
				&  & $0$ & $10$ & $20$ & $30$ & $40$ & $50$ & $60$ & $70$ & $80$ & $90$ & $100$\\
				\cmidrule(lr){3-13}
				& $10$ & \heatmap{4.9} & \heatmap{9.1} & \heatmap{10.0} & \heatmap{12.0} & \heatmap{19.4} & \heatmap{28.5} & \heatmap{40.6} & \heatmap{46.9} & \heatmap{50.8} & \heatmap{53.3} & \heatmap{54.7} \\
				
				& $20$ & \heatmap{10.9} & \heatmap{8.1} & \heatmap{14.9} & \heatmap{26.8} & \heatmap{33.8} & \heatmap{45.1} & \heatmap{50.8} & \heatmap{57.1} & \heatmap{59.1} & \heatmap{60.0} & \heatmap{63.7} \\
				
				& $30$ & \heatmap{9.8} & \heatmap{15.2} & \heatmap{23.9} & \heatmap{28.5} & \heatmap{37.5} & \heatmap{47.9} & \heatmap{54.2} & \heatmap{59.0} & \heatmap{60.4} & \heatmap{62.7} & \heatmap{64.8} \\
				
				& $40$ & \heatmap{15.6} & \heatmap{21.3} & \heatmap{28.9} & \heatmap{33.6} & \heatmap{44.5} & \heatmap{48.4} &\heatmap{53.0} & \heatmap{58.5} & \heatmap{63.0} & \heatmap{61.6} & \heatmap{68.1} \\
				
				& $50$ & \heatmap{20.4} & \heatmap{23.8} & \heatmap{30.6} & \heatmap{38.1} & \heatmap{45.0} & \heatmap{53.4} & \heatmap{56.8} & \heatmap{60.4} & \heatmap{64.0} & \heatmap{65.3} & \heatmap{70.0} \\
				
				& $60$ & \heatmap{22.1} & \heatmap{25.3} & \heatmap{30.6} & \heatmap{37.9} & \heatmap{46.5} & \heatmap{52.0} & \heatmap{59.2} & \heatmap{61.6} & \heatmap{65.6} & \heatmap{68.1} & \heatmap{71.1} \\
				
				& $70$ & \heatmap{20.7} & \heatmap{27.4} & \heatmap{36.8} & \heatmap{41.8} & \heatmap{49.3} & \heatmap{52.0} & \heatmap{59.6} & \heatmap{61.1} & \heatmap{65.2} & \heatmap{69.2} & \heatmap{73.3} \\
				
				& $80$ & \heatmap{23.2} & \heatmap{27.5} & \heatmap{35.9} & \heatmap{42.0} & \heatmap{49.4} & \heatmap{54.2} & \heatmap{60.0} & \heatmap{63.5} & \heatmap{67.0} & \heatmap{69.7} & \heatmap{73.3} \\
				
				& $90$ & \heatmap{21.5} & \heatmap{35.0} & \heatmap{33.3} & \heatmap{41.5} & \heatmap{51.1} & \heatmap{55.8} & \heatmap{58.6} & \heatmap{63.0} & \heatmap{67.0} & \heatmap{69.7} & \heatmap{75.1} \\
				
				& $100$ & \heatmap{22.3} & \heatmap{27.1} & \heatmap{34.6} & \heatmap{45.5} & \heatmap{51.1} & \heatmap{56.3} & \heatmap{62.6} & \heatmap{64.1} & \heatmap{67.7} & \heatmap{71.8} & \heatmap{75.2} \\
				
				\bottomrule
			\end{tabularx}
		\end{subtable}\hfill
		\begin{subtable}[t]{0.49\textwidth}
			\caption{Cross-entropy loss}
			\label{tab:results_ODA_2}
			\begin{tabularx}{\linewidth}{p{0.25cm} | c |*{11}{>{\centering\arraybackslash}X}}
				\toprule
				\multirow{11}{*}[-8ex]{\rotatebox{90}{pseudo-label quantity in $\%$}} &  & \multicolumn{11}{c}{pseudo-label quality in $\%$} \\ 
				\cmidrule(lr){3-13}
				&  & $0$ & $10$ & $20$ & $30$ & $40$ & $50$ & $60$ & $70$ & $80$ & $90$ & $100$\\
				\cmidrule(lr){3-13}
				& $10$ & \heatmap{6.1} & \heatmap{10.9} & \heatmap{5.0} & \heatmap{3.8} & \heatmap{2.2} & \heatmap{6.8} & \heatmap{12.3} & \heatmap{26.5} & \heatmap{39.9} & \heatmap{51.2} & \heatmap{62.1} \\
				
				& $20$ & \heatmap{4.5} & \heatmap{8.2} & \heatmap{4.0} & \heatmap{2.8} & \heatmap{3.2} & \heatmap{6.0} & \heatmap{13.5} & \heatmap{26.8} & \heatmap{41.3} & \heatmap{57.3} & \heatmap{70.5} \\
				
				& $30$ & \heatmap{4.3} & \heatmap{8.7} & \heatmap{2.7} & \heatmap{2.3} & \heatmap{2.4} & \heatmap{4.2} & \heatmap{12.2} & \heatmap{28.3} & \heatmap{46.3} & \heatmap{61.2} & \heatmap{73.6} \\
				
				& $40$ & \heatmap{6.2} & \heatmap{6.7} & \heatmap{3.6} & \heatmap{2.0} & \heatmap{2.2} & \heatmap{4.8} & \heatmap{12.7} & \heatmap{29.3} & \heatmap{48.0} & \heatmap{64.7} & \heatmap{75.5} \\
				
				& $50$ & \heatmap{3.7} & \heatmap{4.8} & \heatmap{2.8} & \heatmap{1.6} & \heatmap{2.3} & \heatmap{5.5} & \heatmap{13.4} & \heatmap{30.1} & \heatmap{48.8} & \heatmap{63.9} & \heatmap{76.1} \\
				
				& $60$ & \heatmap{5.5} & \heatmap{4.8} & \heatmap{2.2} & \heatmap{1.7} & \heatmap{2.6} & \heatmap{6.3} & \heatmap{16.4} & \heatmap{31.2} & \heatmap{49.4} & \heatmap{65.8} & \heatmap{76.4} \\
				
				& $70$ & \heatmap{4.9} & \heatmap{3.4} & \heatmap{2.2} & \heatmap{1.8} & \heatmap{3.2} & \heatmap{7.1} & \heatmap{16.1} & \heatmap{32.0} & \heatmap{50.3} & \heatmap{66.5} & \heatmap{76.7} \\
				
				& $80$ & \heatmap{4.3} & \heatmap{3.1} & \heatmap{2.0} & \heatmap{2.1} & \heatmap{3.7} & \heatmap{8.2} & \heatmap{17.3} & \heatmap{32.2} & \heatmap{49.9} & \heatmap{66.4} & \heatmap{76.8} \\
				
				& $90$ & \heatmap{4.4} & \heatmap{3.1} & \heatmap{2.0} & \heatmap{2.3} & \heatmap{3.9} & \heatmap{8.8} & \heatmap{18.7} & \heatmap{33.0} & \heatmap{50.2} & \heatmap{66.5} & \heatmap{76.8} \\
				
				& $100$ & \heatmap{4.9} & \heatmap{3.1} & \heatmap{2.0} & \heatmap{2.5} & \heatmap{4.6} & \heatmap{10.1} & \heatmap{19.9} & \heatmap{33.4} & \heatmap{50.3} & \heatmap{66.7} & \heatmap{76.8} \\
				
				\bottomrule
			\end{tabularx}
		\end{subtable}
	\end{table*}
	
	\begin{table*}[t!]
		\centering
		\caption{H-scores in $\%$ for the OPDA scenario. The source-only baseline performance is $26.9\%$.}
		\label{tab:results_OPDA}

		\setlength{\tabcolsep}{2.5pt}
		\renewcommand{\arraystretch}{1.2}
		
		\newcommand{\heatmap}[1]{%
			\begingroup
			\def\val{#1} 
			\ifdim\val pt<26.9pt \cellcolor{red!80}#1%
			\else\ifdim\val pt<36.9pt \cellcolor{orange!60}#1%
			\else\ifdim\val pt<46.9pt \cellcolor{yellow!60}#1%
			\else\ifdim\val pt<56.9pt \cellcolor{GreenYellow!80}#1%
			\else\ifdim\val pt<66.9pt \cellcolor{green!80}#1%
			\else \cellcolor{Green!80}#1%
			\fi\fi\fi\fi\fi
			\endgroup
		}
		
		\begin{subtable}[t]{0.49\textwidth}
			\caption{Contrastive loss}
			\label{tab:results_OPDA_1}
			\begin{tabularx}{\linewidth}{p{0.25cm} | c |*{11}{>{\centering\arraybackslash}X}}
				\toprule
				\multirow{11}{*}[-8ex]{\rotatebox{90}{pseudo-label quantity in $\%$}} &  & \multicolumn{11}{c}{pseudo-label quality in $\%$} \\ 
				\cmidrule(lr){3-13}
				&  & $0$ & $10$ & $20$ & $30$ & $40$ & $50$ & $60$ & $70$ & $80$ & $90$ & $100$\\
				\cmidrule(lr){3-13}
				& $10$ & \heatmap{4.3} & \heatmap{9.7} & \heatmap{8.0} & \heatmap{13.0} & \heatmap{13.4} & \heatmap{30.6} & \heatmap{49.8} & \heatmap{51.6} & \heatmap{48.4} & \heatmap{55.7} & \heatmap{58.2} \\
				
				& $20$ & \heatmap{10.0} & \heatmap{18.2} & \heatmap{11.8} & \heatmap{27.2} & \heatmap{32.6} & \heatmap{40.4} & \heatmap{48.0} & \heatmap{50.1} & \heatmap{53.0} & \heatmap{56.7} & \heatmap{60.7} \\
				
				& $30$ & \heatmap{9.6} & \heatmap{20.6} & \heatmap{13.5} & \heatmap{27.9} & \heatmap{34.2} & \heatmap{39.1} & \heatmap{43.3} & \heatmap{54.1} & \heatmap{59.4} & \heatmap{62.5} & \heatmap{65.1} \\
				
				& $40$ & \heatmap{17.7} & \heatmap{20.8} & \heatmap{29.6} & \heatmap{29.6} & \heatmap{35.9} & \heatmap{45.8} & \heatmap{48.8} & \heatmap{53.1} & \heatmap{61.3} & \heatmap{64.7} & \heatmap{68.1} \\
				
				& $50$ & \heatmap{11.2} & \heatmap{20.0} & \heatmap{25.2} & \heatmap{32.7} & \heatmap{39.0} & \heatmap{46.8} & \heatmap{53.2} & \heatmap{55.6} & \heatmap{65.2} & \heatmap{66.0} & \heatmap{68.3} \\
				
				& $60$ & \heatmap{14.8} & \heatmap{24.2} & \heatmap{32.3} & \heatmap{32.9} & \heatmap{41.9} & \heatmap{44.7} & \heatmap{49.2} & \heatmap{58.3} & \heatmap{63.8} & \heatmap{65.3} & \heatmap{68.4} \\
				
				& $70$ & \heatmap{15.2} & \heatmap{29.0} & \heatmap{29.9} & \heatmap{37.6} & \heatmap{41.7} & \heatmap{47.4} & \heatmap{52.5} & \heatmap{56.0} & \heatmap{64.3} & \heatmap{67.9} & \heatmap{70.2} \\
				
				& $80$ & \heatmap{27.8} & \heatmap{26.2} & \heatmap{34.0} & \heatmap{35.5} & \heatmap{40.7} & \heatmap{45.4} & \heatmap{52.3} & \heatmap{60.0} & \heatmap{64.6} & \heatmap{68.7} & \heatmap{69.9} \\
				
				& $90$ & \heatmap{10.9} & \heatmap{30.7} & \heatmap{35.3} & \heatmap{37.7} & \heatmap{38.9} & \heatmap{45.2} & \heatmap{55.1} & \heatmap{59.4} & \heatmap{64.0} & \heatmap{67.9} & \heatmap{68.9} \\
				
				& $100$ & \heatmap{23.6} & \heatmap{28.3} & \heatmap{34.3} & \heatmap{36.9} & \heatmap{44.3} & \heatmap{48.0} & \heatmap{53.4} & \heatmap{58.6} & \heatmap{65.0} & \heatmap{68.2} & \heatmap{71.1} \\
				
				\bottomrule
			\end{tabularx}
		\end{subtable}\hfill
		\begin{subtable}[t]{0.49\textwidth}
			\caption{Cross-entropy loss}
			\label{tab:results_OPDA_2}
			\begin{tabularx}{\linewidth}{p{0.25cm} | c |*{11}{>{\centering\arraybackslash}X}}
				\toprule
				\multirow{11}{*}[-8ex]{\rotatebox{90}{pseudo-label quantity in $\%$}} &  & \multicolumn{11}{c}{pseudo-label quality in $\%$} \\ 
				\cmidrule(lr){3-13}
				&  & $0$ & $10$ & $20$ & $30$ & $40$ & $50$ & $60$ & $70$ & $80$ & $90$ & $100$\\
				\cmidrule(lr){3-13}
				& $10$ & \heatmap{0.4} & \heatmap{1.7} & \heatmap{1.4} & \heatmap{2.0} & \heatmap{4.5} & \heatmap{13.7} & \heatmap{24.8} & \heatmap{41.2} & \heatmap{55.6} & \heatmap{67.7} & \heatmap{76.1} \\
				
				& $20$ & \heatmap{5.2} & \heatmap{3.9} & \heatmap{1.5} & \heatmap{2.1} & \heatmap{4.9} & \heatmap{13.0} & \heatmap{26.0} & \heatmap{44.4} & \heatmap{60.2} & \heatmap{73.8} & \heatmap{80.0} \\
				
				& $30$ & \heatmap{3.4} & \heatmap{1.2} & \heatmap{1.3} & \heatmap{2.0} & \heatmap{5.0} & \heatmap{12.7} & \heatmap{27.5} & \heatmap{44.5} & \heatmap{63.1} & \heatmap{75.1} & \heatmap{81.8} \\
				
				& $40$ & \heatmap{2.8} & \heatmap{3.4} & \heatmap{1.8} & \heatmap{2.0} & \heatmap{5.9} & \heatmap{14.4} & \heatmap{28.0} & \heatmap{45.4} & \heatmap{62.9} & \heatmap{75.6} & \heatmap{82.1} \\
				
				& $50$ & \heatmap{2.9} & \heatmap{1.9} & \heatmap{2.2} & \heatmap{3.1} & \heatmap{8.3} & \heatmap{17.3} & \heatmap{29.0} & \heatmap{46.2} & \heatmap{63.8} & \heatmap{76.3} & \heatmap{82.3} \\
				
				& $60$ & \heatmap{2.3} & \heatmap{2.4} & \heatmap{2.3} & \heatmap{4.3} & \heatmap{10.8} & \heatmap{18.4} & \heatmap{29.7} & \heatmap{46.4} & \heatmap{63.3} & \heatmap{76.6} & \heatmap{82.5} \\
				
				& $70$ & \heatmap{3.1} & \heatmap{2.4} & \heatmap{3.1} & \heatmap{5.6} & \heatmap{12.2} & \heatmap{19.3} & \heatmap{31.1} & \heatmap{46.2} & \heatmap{64.4} & \heatmap{76.9} & \heatmap{82.3} \\
				
				& $80$ & \heatmap{2.9} & \heatmap{2.3} & \heatmap{2.9} & \heatmap{7.0} & \heatmap{13.1} & \heatmap{20.7} & \heatmap{31.4} & \heatmap{47.2} & \heatmap{64.3} & \heatmap{76.6} & \heatmap{82.4} \\
				
				& $90$ & \heatmap{2.9} & \heatmap{1.9} & \heatmap{3.0} & \heatmap{8.6} & \heatmap{13.9} & \heatmap{21.3} & \heatmap{31.7} & \heatmap{46.9} & \heatmap{64.2} & \heatmap{76.6} & \heatmap{82.3} \\
				
				& $100$ & \heatmap{2.5} & \heatmap{1.4} & \heatmap{2.9} & \heatmap{8.9} & \heatmap{14.8} & \heatmap{21.5} & \heatmap{32.0} & \heatmap{46.4} & \heatmap{63.7} & \heatmap{76.5} & \heatmap{82.5} \\
				
				\bottomrule
			\end{tabularx}
		\end{subtable}
	\end{table*}
	
	\section{Discussion}
	The experiments provide valuable insights into the relationship between pseudo-labeling and adaptation outcomes in online SF-UniDA. In the following, we discuss the key findings by answering the four formulated research questions.
	\subsection{Upper Bound of Adaptation (Q1)}
	As expected, the best results are achieved with perfect pseudo-labeling (100\% pseudo-label quality and quantity). These results significantly surpass the current state-of-the-art achieved by \cite{GMM}, which yields an accuracy of $41.2\%$ for PDA, an H-score of $59.9\%$ for ODA, and an H-score of $60.3\%$ for OPDA. Specifically, our simulation demonstrates that perfect pseudo-labeling achieves an accuracy of 75.6\% for PDA, an H-score of $74.8\%$ for ODA, and an H-score of $71.1\%$ for OPDA using the contrastive loss-based approach. Even more strikingly, the CE loss-based approach reaches $91.9\%$ for PDA, $76.7\%$ for ODA, and $82.5\%$ for OPDA with ideal pseudo-labeling. Hence, these results reveal a substantial gap between the upper bound of adaptation performance and the current state-of-the-art, highlighting the significant potential for improvement in existing methods. This margin underscores the critical role of pseudo-labeling accuracy and reliability in achieving optimal adaptation outcomes, suggesting that advancements in pseudo-labeling strategies could lead to substantial performance gains across all scenarios.
	
	\subsection{Pseudo-Label Requirements for Effective Adaptation (Q2)}
	The minimum pseudo-labeling requirements for effective adaptation depend heavily on the chosen loss function. Our results show that for the contrastive loss-based approach, adaptation remains effective even with a pseudo-label accuracy as low as $30\%$, as long as the pseudo-label quantity is sufficiently high ($\geq40\%$). In contrast, when using CE loss, adaptation performance deteriorates much more rapidly as pseudo-label quality decreases, resulting in significantly higher minimum requirements. Specifically, a pseudo-label accuracy of at least $70\%$ is necessary to achieve improvements over the source-only model across all category shift scenarios.
	
	\subsection{Comparison of Contrastive and Cross-Entropy Loss (Q3)}
	As already observed previously, the contrastive loss-based approach is significantly more robust to erroneous pseudo-labels while the CE loss is able to achieve substantially better results if the pseudo-label quality approaches $100\%$.
	
	In a closer analysis, we find that the poor performance of the CE loss with low pseudo-label quality stems from the fact that it classifies nearly all samples as unknown. This behavior is not solely caused by the pseudo-labeling process misclassifying known classes as unknown, as setting $\alpha=0$ (i.e., disallowing falsely unknown pseudo-labels) results in only small improvements. Instead, the issue appears to be inherent to the nature of the CE loss. We hypothesize that, due to its direct enforcement of desired model outputs, even small amounts of incorrect pseudo-labels can confuse the model, leading to many uncertain predictions that are subsequently interpreted as unknown.
	
	In contrast, the contrastive loss operates on similarity constraints in the feature space. This allows to maintain better performance in cases where pseudo-labeling is imperfect, as it focuses more on relative relationships between samples rather than explicit class assignments. Consequently, the contrastive loss provides a more stable adaptation process, making it the preferable choice when dealing with unreliable pseudo-labels.
	
	\subsection{Importance of Pseudo-Label Quality vs. Quantity (Q4)}
	The results confirm the intuition that pseudo-label quality plays a more critical role than quantity for effective adaptation. Across all scenarios, the adaptation performance generally declines much more rapidly as pseudo-label quality decreases compared to the impact of reduced pseudo-label quantity. This trend indicates that an adaptation approach focused on high-confidence pseudo-labels—despite using fewer samples—yields in general more effective results than one that relies on a larger number of uncertain pseudo-labels.
	
	We assume that the reason for this observation is that incorrect pseudo-labels not only fail to improve the model but actively degrade its performance by steering adaptation in the wrong direction. Intuitively, it is preferable to maintain the model unchanged rather than undergoing such negative adaptation. Accordingly, maximizing pseudo-label quality is the key priority, whereas quantity remains of secondary importance.
	
	\subsection{Further observations}
	In addition to addressing the core research questions, our analysis also reveals a noteworthy difference regarding the sensitivity to the pseudo-label quality between PDA and the other two category shifts. In the PDA scenario, which involves only known classes, adaptation performance is much more sensitive to pseudo-label quality. For example, using the contrastive loss-based approach, the performance difference between the best results for perfect pseudo-labeling quality ($100\%$) and the best result for a pseudo-labeling quality of $90\%$ is more than $17\%$, while this difference is below $4\%$ for both ODA and OPDA. This discrepancy most likely arises because, in PDA, all predictions of samples as unknown directly correspond to prediction errors, as the target domain contains no unknown classes. In contrast, in ODA and OPDA, predicting samples as unknown is a necessary and desired behavior which needs to be encouraged by the adaptation as the source model is only trained on known classes. This fundamental difference highlights one of the main challenges of UniDA, where the adaptation method must perform well for all three scenarios without having prior knowledge. By following the tendency of the pseudo-labeling of \cite{COMET} and \cite{GMM} to predominantly misclassifying samples of known classes as unknown, this effect is even enhanced in the given simulation in favor of the ODA and OPDA scenarios. We could verify this statement by repeating one arbitrary experiment (quality $50\%$, quantity $100\%$) using $\alpha=0$ for the pseudo-label simulation, meaning that no falsely unknown pseudo-labels occur. The result for PDA is with $41.4\%$ more than $22\%$ higher than the result of the original simulation. This insight emphasizes the need of SF-UniDA methods to balance the individual requirements of all three scenarios.
	
	\section{Limitations}
	Since our pseudo-label simulation was designed to analyze adaptation performance under a fixed, predefined global pseudo-label quality and quantity, it does not fully capture the dynamics of real pseudo-labeling during the adaptation process. In practice, adaptation itself influences pseudo-labeling, typically leading to an increase in batchwise pseudo-label quality and quantity over time.
	
	In this sense, the choice of the loss function not only directly affects the adaptation outcome but also has an indirect influence by guiding the improvement of pseudo-labeling throughout the adaptation process. For instance, a contrastive loss promotes clustering of known classes in the feature space, which aligns well with GMM-based pseudo-labeling that assumes class distributions follow Gaussian clusters. However, our simulation, relying on synthetic pseudo-labeling, could not capture this interaction between the loss function and the evolution of pseudo-labeling. Therefore, in addition to the insights we presented in response to research question Q3, this effect should be considered when selecting a loss function and deserves further investigation.
	
	\section{Conclusion}
	In conclusion, this work underscores the critical role of pseudo-labeling in online SF-UniDA and provides actionable insights for improving adaptation performance. By addressing the research questions, we contribute to a deeper understanding of the trade-offs between pseudo-label quality and quantity, the robustness of different loss functions, and the maximum achievable adaptation outcomes. The results highlight a significant gap between current state-of-the-art methods and the upper bound of performance achievable with perfect pseudo-labeling, emphasizing the need for further advancements in pseudo-labeling strategies. At the same time, the detailed insights and systematic analysis provided in this work offer valuable guidance for future research, paving the way for more effective approaches to narrow this gap and enhance adaptation performance in real-world applications.

	\bibliographystyle{IEEEtran}
	
	\bibliography{IEEEabrv, refs}

\begin{thebibliography}{10}
\providecommand{\url}[1]{#1}
\csname url@samestyle\endcsname
\providecommand{\newblock}{\relax}
\providecommand{\bibinfo}[2]{#2}
\providecommand{\BIBentrySTDinterwordspacing}{\spaceskip=0pt\relax}
\providecommand{\BIBentryALTinterwordstretchfactor}{4}
\providecommand{\BIBentryALTinterwordspacing}{\spaceskip=\fontdimen2\font plus
\BIBentryALTinterwordstretchfactor\fontdimen3\font minus
  \fontdimen4\font\relax}
\providecommand{\BIBforeignlanguage}[2]{{%
\expandafter\ifx\csname l@#1\endcsname\relax
\typeout{** WARNING: IEEEtran.bst: No hyphenation pattern has been}%
\typeout{** loaded for the language `#1'. Using the pattern for}%
\typeout{** the default language instead.}%
\else
\language=\csname l@#1\endcsname
\fi
#2}}
\providecommand{\BIBdecl}{\relax}
\BIBdecl

\bibitem{GLC}
S.~Qu, T.~Zou, F.~R{\"o}hrbein, C.~Lu, G.~Chen, D.~Tao, and C.~Jiang,
  ``Upcycling models under domain and category shift,'' in \emph{Proceedings of
  the IEEE/CVF Conference on Computer Vision and Pattern Recognition}, 2023,
  pp. 20\,019--20\,028.

\bibitem{COMET}
P.~Schlachter and B.~Yang, ``Comet: Contrastive mean teacher for online
  source-free universal domain adaptation,'' in \emph{2024 International Joint
  Conference on Neural Networks (IJCNN)}, 2024, pp. 1--9.

\bibitem{GMM}
P.~Schlachter, S.~Wagner, and B.~Yang, ``Memory-efficient pseudo-labeling for
  online source-free universal domain adaptation using a gaussian mixture
  model,'' in \emph{2025 IEEE/CVF Winter Conference on Applications of Computer
  Vision (WACV)}, 2025.

\bibitem{GLC++}
S.~Qu, T.~Zou, F.~R{\"o}hrbein, C.~Lu, G.~Chen, D.~Tao, and C.~Jiang, ``Glc++:
  Source-free universal domain adaptation through global-local clustering and
  contrastive affinity learning,'' \emph{arXiv preprint arXiv:2403.14410},
  2024.

\bibitem{LEAD}
S.~Qu, T.~Zou, L.~He, F.~Röhrbein, A.~Knoll, G.~Chen, and C.~Jiang, ``Lead:
  Learning decomposition for source-free universal domain adaptation,'' 2024.

\bibitem{feng2021open}
Z.~Feng, C.~Xu, and D.~Tao, ``Open-set hypothesis transfer with semantic
  consistency,'' \emph{IEEE Transactions on Image Processing}, vol.~30, pp.
  6473--6484, 2021.

\bibitem{universal_sf_da}
J.~N. Kundu, N.~Venkat, R.~V. Babu \emph{et~al.}, ``Universal source-free
  domain adaptation,'' in \emph{Proceedings of the IEEE/CVF Conference on
  Computer Vision and Pattern Recognition}, 2020, pp. 4544--4553.

\bibitem{kundu2020towards}
J.~N. Kundu, N.~Venkat, A.~Revanur, R.~V. Babu \emph{et~al.}, ``Towards
  inheritable models for open-set domain adaptation,'' in \emph{Proceedings of
  the IEEE/CVF conference on computer vision and pattern recognition}, 2020,
  pp. 12\,376--12\,385.

\bibitem{umad}
J.~Liang, D.~Hu, J.~Feng, and R.~He, ``Umad: Universal model adaptation under
  domain and category shift,'' 2021.

\bibitem{coca}
X.~Liu, Y.~Zhou, T.~Zhou, C.-M. Feng, and L.~Shao, ``Coca: Classifier-oriented
  calibration via textual prototype for source-free universal domain
  adaptation,'' 2024.

\bibitem{visda}
X.~Peng, B.~Usman, N.~Kaushik, D.~Wang, J.~Hoffman, and K.~Saenko, ``Visda: A
  synthetic-to-real benchmark for visual domain adaptation,'' in
  \emph{Proceedings of the IEEE Conference on Computer Vision and Pattern
  Recognition Workshops}, 2018, pp. 2021--2026.

\bibitem{coco}
T.-Y. Lin, M.~Maire, S.~Belongie, J.~Hays, P.~Perona, D.~Ramanan,
  P.~Doll{\'a}r, and C.~L. Zitnick, ``Microsoft coco: Common objects in
  context,'' in \emph{Computer Vision--ECCV 2014: 13th European Conference,
  Zurich, Switzerland, September 6-12, 2014, Proceedings, Part V 13}.\hskip 1em
  plus 0.5em minus 0.4em\relax Springer, 2014, pp. 740--755.

\bibitem{resnet}
K.~He, X.~Zhang, S.~Ren, and J.~Sun, ``Deep residual learning for image
  recognition,'' in \emph{Proceedings of the IEEE conference on computer vision
  and pattern recognition}, 2016, pp. 770--778.

\end{thebibliography}
	
\end{document}